%% file: arxiv.tex
\documentclass{article}

    \usepackage[preprint]{neurips_2025}

\usepackage[utf8]{inputenc} 
\usepackage[T1]{fontenc}    
\usepackage{hyperref}       
\usepackage{url}            
\usepackage{booktabs}       
\usepackage{amsfonts}       
\usepackage{nicefrac}       
\usepackage{microtype}      
\usepackage{xcolor}         
\usepackage{color}
\usepackage{colortbl}
\usepackage{graphicx}
\usepackage{amsmath}
\usepackage{amssymb}
\usepackage{xspace}
\usepackage{caption}
\usepackage{adjustbox}  
\usepackage{subcaption}
\usepackage{fontawesome}

\newcommand{\method}[0]{\textsc{VideoMolmo}\xspace}

\title{\method: Spatio-Temporal Grounding \\ Meets Pointing}

\author{%
Ghazi Shazan Ahmad$^{1*}$ \quad Ahmed Heakl$^{1*}$ \quad Hanan Gani$^{1}$ \quad  \\ \textbf{Abdelrahman Shaker}$^1$ \quad  \textbf{Zhiqiang Shen}$^1$ \quad \textbf{Fahad Shahbaz Khan}$^{1,4}$ \quad 
\textbf{Salman Khan}$^{1,5}$ \\ \\
$^1$Mohamed Bin Zayed University of Artificial Intelligence
\quad $^4$Link{\"o}ping University \\
$^5$Australian National University 
\\
Correspondence: \texttt{\{ghazi.ahmad, ahmed.heakl, hanan.ghani\} @mbzuai.ac.ae}\\
\\
\faGlobe\ \texttt{https://mbzuai-oryx.github.io/VideoMolmo}
    \vspace{-0.5em}
}

\begin{document}

\maketitle
\def\thefootnote{*}\footnotetext{Equal technical contribution}
\begin{figure}[h]
    \centering
    \includegraphics[width=\textwidth]{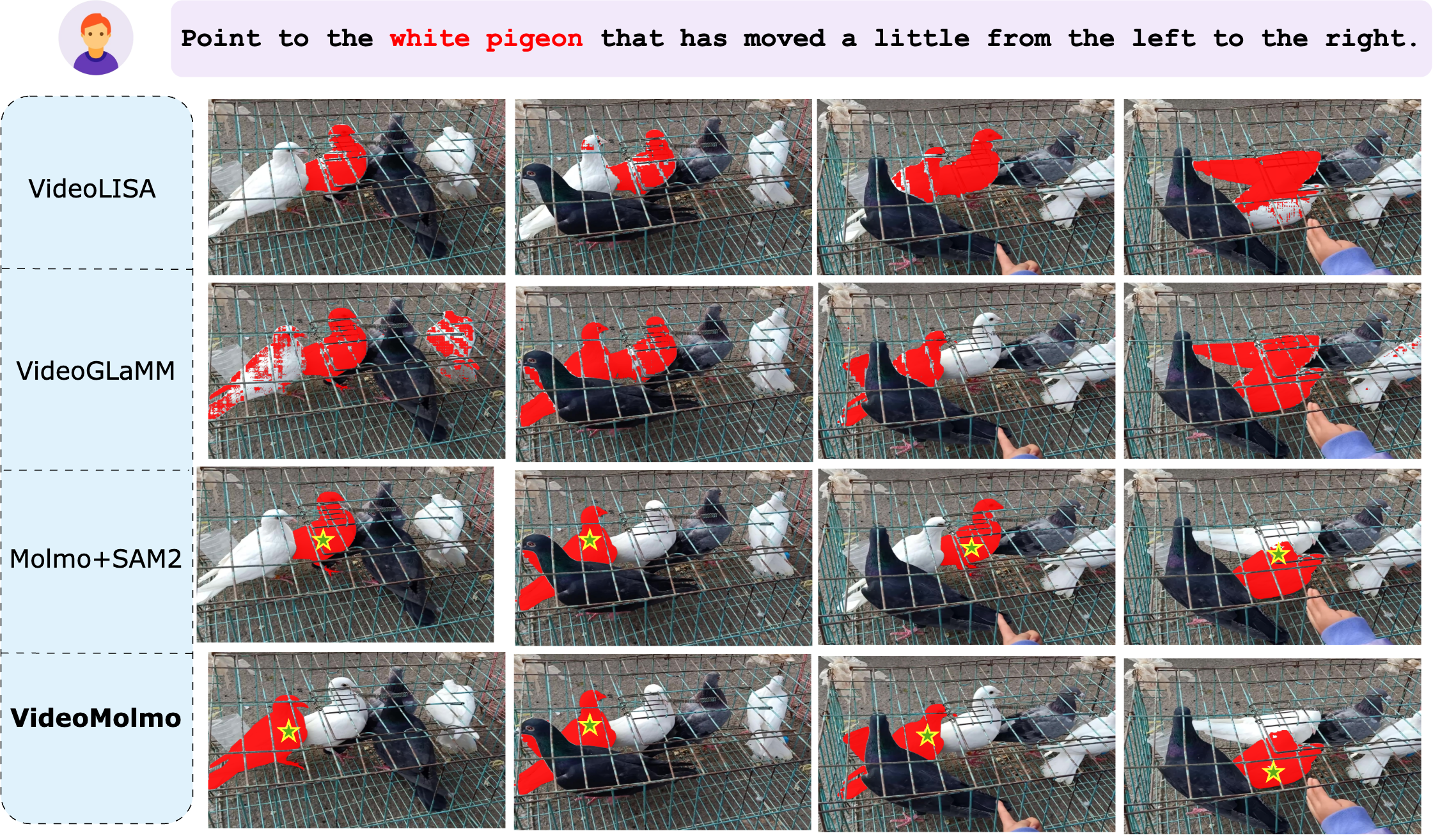}
    \caption{Given complex referring expressions in natural language, \method demonstrates improved spatio-temporal reasoning in visual grounding. 
    By decomposing the visual grounding task into sequential steps—pointing (denoted by star) followed by generating masks (in red) -\method produces more accurate and coherent segmentation masks compared to prior approaches.}
\label{fig:teaser}
\end{figure}

\begin{abstract}
\input{sections/0-abstract}
\end{abstract}

\input{sections/1-introduction}
\vspace{-0.75em}
\input{sections/2-relatedwork}

\vspace{-0.2em}
\input{sections/3-method}

\input{sections/4-dataset}
\input{sections/5-experiments}
\input{sections/7-conclusion}

\bibliographystyle{plain}
\bibliography{refs}

\newpage
\clearpage
\input{sections/99-appendix}


\end{document}

%% file: sections/0-abstract.tex
Spatio-temporal localization is vital for precise interactions across diverse domains, from biological research to autonomous navigation and interactive interfaces. Current video-based approaches, while proficient in tracking, lack the sophisticated reasoning capabilities of large language models, limiting their contextual understanding and generalization. We introduce \method, a large multimodal model tailored for fine-grained spatio-temporal pointing conditioned on textual descriptions. Building upon the Molmo architecture, \method incorporates a temporal module utilizing an attention mechanism to condition each frame on preceding frames, ensuring temporal consistency. Additionally, our novel temporal mask fusion pipeline employs SAM2 for bidirectional point propagation, significantly enhancing coherence across video sequences. This two‐step decomposition i.e., first using the LLM to generate precise pointing coordinates, then relying on a sequential mask‐fusion module to produce coherent segmentation, not only simplifies the task for the language model but also enhances interpretability. Due to the lack of suitable datasets, we curate a comprehensive dataset comprising 72k video-caption pairs annotated with 100k object points. To evaluate the generalization of VideoMolmo, we introduce VPoS-Bench, a challenging out-of-distribution benchmark spanning five real-world scenarios: Cell Tracking, Egocentric Vision, Autonomous Driving, Video-GUI Interaction, and Robotics. We also evaluate our model on Referring Video Object Segmentation (Refer-VOS) and Reasoning VOS tasks. In comparison to existing models, \method substantially improves spatio-temporal pointing accuracy and reasoning capability. Our code and models are publicly available at \textcolor{magenta}{\url{https://github.com/mbzuai-oryx/VideoMolmo}}.

%% file: sections/1-introduction.tex
\section{Introduction}
Precise spatio‐temporal localization underpins a wide range of applications, enabling fine-grained interactions and analysis in dynamic environments. For instance, accurately tracking cell nuclei movement is essential in biological research, aiding in understanding developmental processes and disease progression~\cite{meijering2012tracking}. Similarly, autonomous vehicles rely on continuous tracking of pedestrians, vehicles and traffic lights to ensure safe navigation; robots manipulating objects must precisely localize contact points over time while avoiding collisions~\cite{yuan2024robopoint}. In egocentric videos, captured from wearable, first-person cameras, spatio-temporal localization is key for tasks such as daily activity recognition, assistive technologies for visually impaired users, and modeling hand-object interactions in real-world settings~\cite{grauman2022ego4d}. Counting objects across frames is also vital in scenarios such as surveillance and traffic monitoring, where assessing object quantities informs critical decisions (Fig.~\ref{fig:benchmark-qualitative}).

Prior efforts \cite{nguyen2023type, yang2020grounding, DBLP:journals/prl/ZhaoWWLR23, wang2021towards, 9857151} have explored text-guided tracking and grounding in videos. However, these approaches often lack the reasoning depth offered by large language models (LLMs), limiting their generalization and contextual understanding.
While recent video large multimodal models (LMMs) can generate dense, spatio‐temporally coherent masks~\cite{bai2024one, munasinghe2024videoglamm}, none support free‐form, text‐conditioned pointing in videos.

For instance, as illustrated in Fig.~\ref{fig:teaser}, the instruction is to identify “the white pigeon that has moved slightly from left to right” among five pigeons in close proximity in a video sequence. Solving such a task demands both temporal understanding and fine-grained reasoning to distinguish subtle object movements. Existing methods often struggle in such scenarios, frequently predicting multiple objects or localizing the wrong one, highlighting the limitations of models without integrated reasoning and fine-grained localization capabilities.

Image‐based pointing LMMs have demonstrated strong single‐frame performance~\cite{deitke2024molmo,yuan2024robopoint}, but they cannot model the temporal dynamics essential for video tasks.
To fill this gap, we introduce \method, an LMM that accepts natural‐language queries and produces point‐level predictions for target objects across entire video sequences, ensuring temporal consistency.
In this way, \method decomposes visual grounding in videos into two simpler stages: first, generating precise pointing coordinates via the LLM, then sequentially fusing these points into coherent masks with a dedicated module. This decomposition simplifies the problem for the language model and improves overall interpretability.

\method is built on top of Molmo \cite{deitke2024molmo}, which features a pre-processor that converts the input image into multiscale, multi-crop images, a vision encoder, an LLM decoder and a visual projector that
pools and projects image features into the LLM’s embedding
space. While Molmo processes images independently, we adapt Molmo architecture to process video data in a simplified manner. We first introduce a temporal module that handles the temporal nature of the video data by conditioning each frame on previous frames through an attention mechanism.
Additionally, we propose a novel temporal mask fusion pipeline that leverages SAM2 for bidirectional point propagation, enhancing temporal coherence by utilizing local structural cues embedded in videos.
Since there exists no spatio-temporal pointing dataset to train such systems, we release a comprehensive dataset of 72k video-caption pairs and 100k object points. 
To evaluate the generalization of VideoMolmo, we introduce VPoS-Bench, a challenging out-of-distribution benchmark spanning five real-world scenarios: Cell Tracking, Egocentric Vision, Autonomous Driving, Video-GUI Interaction, and Robotics. We also assess our model on Referring Video Object Segmentation (Ref-VOS) and Reasoning VOS tasks, which require masks as output. For this, we leverage SAM2 to convert the predicted points into segmentation masks and propose a bidirectional temporal mask fusion technique that enhances mask consistency without further training. Experimental results show that VideoMolmo outperforms existing approaches across all benchmarks and task settings, offering a generalizable solution for fine-grained language-guided reasoning in dynamic visual environments. Specifically, on our challenging VPoS- Bench, VideoMolmo exhibits $5.4$ pp average improvement compared to the strongest baseline (Table \ref{table:videomolmo-benchmark}). Similarly, on the challenging referring segmentation on MeViS \cite{MeViS}, VideoMolmo outperforms the strongest baseline by $9.5$ pp (Table \ref{table:ref-seg-merged}), highlighting its effectiveness in visual grounding and reasoning tasks.

%% file: sections/2-relatedwork.tex
\section{Related work}

\noindent \textbf{Video-LMMs.}
Multi-modal LLMs such as \cite{liu2023visual,zhu2023minigpt} have demonstrated notable advancements due to their strong zero-shot abilities, which have been possible due to their training on millions of image-text pairs. Typically, such models project visual information in the latent space of LLM via an encoder and a connector and thus align the information from vision and text modality. The work in LMMs paved the way for the development of Video-LMMs  \cite{2023videochat,damonlpsg2023videollama,lin2023video,maaz2024videogpt+,wang2024internvideo2,zhu2025internvl3,bai2025qwen2,zhang2025videollama}, which unlike image-based LMMs, can reason on dynamic video content. While effective for overall video input comprehension, these methods fell short of fine-grained visual grounding in videos.\\ 
\noindent \textbf{Visual grounding. }
Recent works in Grounded LMMs \cite{rasheed2023glamm} have sparked tremendous interest among the research community. Visual grounding \cite{liu2021dimbert}
seeks to identify the location of nouns or short phrases (such as a man with blue shirt) within an image. These models are trained on a huge dataset of image-caption pairs along with the dense segmentation mask associated with the objects in the caption. \cite{bai2024one,munasinghe2024videoglamm} extended the grounding to video data by releasing a large dataset of  grounded video-QA triplet pairs along with the masks associated with the objects. Training on such huge video grounded datasets allowed for video grounding. On the contrary, our proposed VideoMolmo model and dataset are designed for outputting precise object level points which are crucial for applications such as autonomous driving, counting tasks, robotics etc.\\
\noindent
\textbf{Language-assisted object tracking. } Most text-based tracking methods are limited to tracking a single object only \cite{yang2020grounding,DBLP:journals/prl/ZhaoWWLR23,wang2021towards,9857151}. However, real-world applications can feature multiple object trajectories, making it harder for the single object tracking methods. \cite{nguyen2023type} propose Type-toTrack along with a tracking dataset `GroOT' for multi-object tracking. However, they track objects via bounding boxes and not precise points which limits their applicability. Another work SAM-PT \cite{rajivc2025segment} proposes to use SAM \cite{ravi2024sam} model along with a long-term point tracking mechanism for point-centric interactive video segmentation. However, since their method adapts a 2D model to handle video data, it faces challenges in temporal consistency, especially in the cases of occlusions and fast-moving objects. On the contrary, our proposed \method is trained end-to-end on our training dataset and maintains temporal consistency via a dedicated memory module.

%% file: sections/3-method.tex
\vspace{-0.5em}
\section{\method}
\vspace{-0.5em}
\method is trained end-to-end for spatio-temporal pointing conditioned on textual instructions. It features a visual encoder, a temporal module, an LLM, and a post-processing module. The visual encoder processes the video frames and outputs multi-crop features. To maintain temporal consistency, we introduce a temporal module which employs a cross-attention operation and ensures that the current frame attends to the information in previous frames. The resultant features are then passed to the LLM, which, along with the textual query, processes this information and outputs the points corresponding to the objects. Our post-processing module takes the predictions from \method and uses SAM2 to propagate the points across all video frames bidirectionally. 


\maketitle
\begin{figure}[!t]
    \centering
    \includegraphics[width=1.0\textwidth]{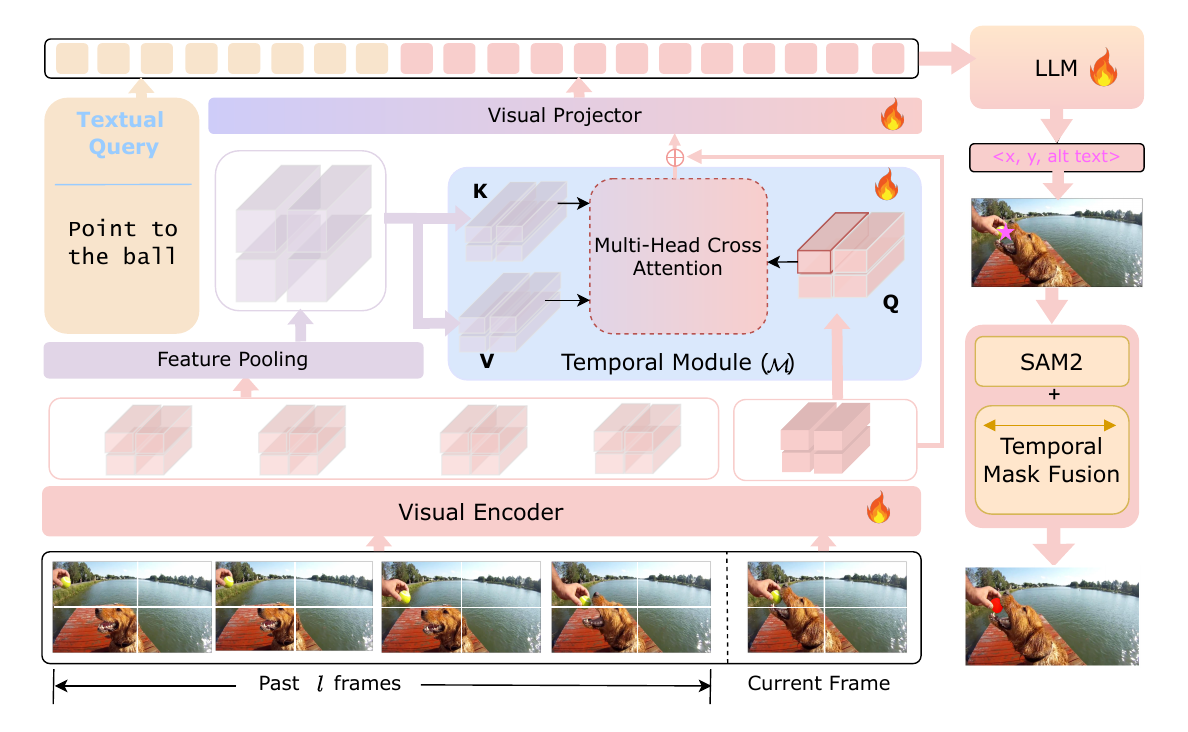}
    \vspace{-2em}
    \caption{\textbf{\method Architecture.} The visual encoder extracts multi-crop features from the current frame and the past $l$ frames. These temporal features provide contextual cues and are processed by the Temporal Module $\mathcal{M}$ via multi-head cross-attention, where the query comes from the current frame, and key and value from the mean of previous frames. The output is fused with the original features to enrich temporal cues while preserving the spatial details of the current frame. The combined visual-textual representations are then passed to the LLM to predict grounded points. These points are converted into masks using our Bidirectional Temporal Mask Fusion module, ensuring temporally consistent pixel-level grounding.}
\label{fig:main_diagram}
\vspace{-1.3em}
\end{figure}

\subsection{Architecture}
\method extends the Molmo architecture~\cite{deitke2024molmo} from static image understanding to spatio-temporal video grounding. The model consists of four end-to-end trainable components: (1) a visual encoder, (2) a temporal module, (3) visual projector (4) a decoder-only large language model (LLM); and a post-processing module (see Fig.~\ref{fig:main_diagram}).

We represent an input video as $V \in \mathbb{R}^{|\mathcal{T}|\times H \times W \times C}$ where $H$, $W$, and $C$ denote the height, width, and number of channels respectively of the frames and $|\mathcal{T}|$ are the number of frames in the video. For a frame $\mathcal{T}_i \in V$, we generate $N$ spatial overlapping multi-crops to capture both global context and fine-grained details in the frame. The multi-crops of the frame are processed independently using a visual encoder $\mathcal{F}$. Following \cite{deitke2024molmo}, we build patch features for each crop by concatenating features from third-to-last and tenth-to-last ViT layers. The features extracted from each crop of frame $\mathcal{T}_i$ are denoted by $f_{\mathcal{T}_i}^j \in \mathbb{R}^{P\times D}$, where $j$ denotes the crop index, $P$ is the number of patches and $D$ is the latent dimension of the ViT.
Since the components in our model are based on 2D modules, we inject temporal information through a dedicated temporal module denoted as $\mathcal{M}$ (Sec. ~\ref{temporal-module}). 
For each frame $\mathcal{T}_i$, we compute a temporal context feature by aggregating features from the $l$ most recent frames $\{\mathcal{T}_{i-l}, \mathcal{T}_{i - l - 1}, \ldots, \mathcal{T}_{i-1}\}$ such that ($l < i$) and denote their mean as $f_{\mathcal{T}_{i-}^*}$. The pooled context features and the current frame features are then passed into the temporal module.
The temporally enriched features predicted by the temporal module $\mathcal{M}$ correspond to $2 \times 2$ patch window features extracted from all the patches of the crops. These window-level features are added to the current frame features by reshaping $f_{\mathcal{T}_{i}}$ from $\mathbb{R}^{N \times P \times D}$ to $\mathbb{R}^{N.\frac{P}{4} \times 4 \times D}$. 
\begin{equation}
    \hat{f}_{\mathcal{T}_i} = f_{\mathcal{T}_{i}} + \mathcal{M}\big(f_{\mathcal{T}_{i}},f_{\mathcal{T}_{i-}^*}\big)
\end{equation}
These resultant features $\hat{f}_{\mathcal{T}_{i}}$ are then aggregated into a single frame-level representation using an attention pooling mechanism.
The pooled feature $\hat{f}_{\mathcal{T}_i}$ is then projected into the embedding space of the language model using a learnable projection layer $\mathcal{P}$. It is then concatenated with the tokenized query $q$ and passed to the decoder-only LLM to obtain the final output text $\mathbf{p}$ which contains 
 the coordinate $\{(x, y)\}^{\mathcal{O}}$ where $\mathcal{O}$ is the number of objects grounded in the frame.
\begin{equation}
\mathbf{p} = \texttt{LLM}\left(\left[\mathcal{P}(\hat{f}_{\mathcal{T}_i});\ q\right]\right)
\end{equation}
The model is trained end-to-end by minimizing the cross-entropy loss \(\mathcal{L}_{\mathrm{CE}}\) between the predicted text \(\mathbf{p}\) and the ground-truth one-hot labels \(\mathbf{p}^*\) in an auto-regressive manner:
\begin{equation}
\mathcal{L}_{\mathrm{CE}}
=   -\mathbf{p}^* \cdot \log(\mathbf{p})    
\end{equation}

\subsection{Temporal Module}
\label{temporal-module}
The original MoLMo architecture~\cite{deitke2024molmo} was developed for static images and cannot model the temporal dynamics inherent in video data. To address this, we introduce a dedicated \textbf{temporal module} $\mathcal{M}$ that infuses each frame with temporal context from the preceding $l$ frames \cite{nguyen2023type,lai2020mast}.
For each crop $j$ of frame $\mathcal{T}_i$, patch features $f_{\mathcal{T}_i}^j \in \mathbb{R}^{P \times D}$ are extracted and reshaped into non‐overlapping $2\times 2$ windows of four vectors, $f_{\mathcal{T}_i}^j = \{\,f_{\mathcal{T}_{i}}^{j, p} \in \mathbb{R}^{4\times D}\}_{p=1}^{P/4}$. We then flatten the window features across all $N$ crops to obtain resized fetures called window vectors $f_{\mathcal{T}_i}
$ and $
f_{\mathcal{T}_{i-}^*} $ such that both lie in $\mathbb{R}^{(N\cdot P/4)\times 4\times D}$ for both the current features and the context features.
To capture fine-grained temporal correspondences, we apply multi-head cross-attention (MHCA) over each patch window, where the query comes from current frame window vector $f_{\mathcal{T}_{i}}$ , and key and value come from  vector $f_{\mathcal{T}_{i-}^*}$ such that $ \texttt{MHCA}\left(
     f_{\mathcal{T}_i},
    \;f_{\mathcal{T}_{i-}^*},\;f_{\mathcal{T}_{i-}^*}
\right)$ denotes the final attended window features.
By restricting attention to small $2\times2$ neighborhoods, the module selectively refines each local region of the current frame using only the most relevant fine‐grained patches from its temporal context. This targeted cross‐attention both preserves spatial locality and amplifies subtle motion cues that would be lost in coarse global pooling.

\subsection{Bidirectional Temporal Mask Fusion}
\label{sec:post-processing}
While our method predicts grounded point coordinates corresponding to objects mentioned in the textual query, most existing evaluation protocols are designed to operate on segmentation masks. Therefore, for compatibility and consistent evaluation, we introduce Bidirectional Temporal Mask Fusion, a novel post-processing technique that leverages SAM2 to convert points to dense masks. Specifically, we use the predicted points as prompts to SAM2 \cite{ravi2024sam} to obtain the segmentation masks of the objects.
Due to computational constraints, we avoid dense frame-level inference across the entire video. Instead, we sparsely sample frames at a sampling rate $k$, and generate point predictions only on these sampled frames. Let the two consecutive sampled frames be denoted as $\mathcal{T}_i$ and $\mathcal{T}_{i+k}$. The corresponding point predictions are transformed into segmentation masks $m_i$ and $m_{i+k}$ using SAM2.
For frames lying between the two sampled frames (i.e., $\mathcal{T}_{i+n}$ for $0 < n < k$), we propagate the neighboring masks $m_i$ and $m_{i+k}$ bidirectionally to estimate the intermediate masks. The leftward propagation from $m_i$ and the rightward propagation from $m_{i+k}$ are performed using SAM2’s temporal propagation module, defined as,
\begin{equation}
\begin{aligned}
    \widehat{m}_{i+n}^{\rightarrow} &= \texttt{Prop}^{\rightarrow}(\{\mathcal{T}_i, m_i\}, \mathcal{T}_{i+n}) \\
    \widehat{m}_{i+n}^{\leftarrow} &= \texttt{Prop}^{\leftarrow}(\{\mathcal{T}_{i+k}, m_{i+k}\}, \mathcal{T}_{i+n}),
\end{aligned}
\end{equation}
where $\widehat{m}_{i+n}^{\rightarrow}$ and $\widehat{m}_{i+n}^{\leftarrow}$ denote the masks propagated from the left and right directions, respectively.
To reconcile the two propagated masks at frame $\mathcal{T}_{i+n}$, we compute their Intersection over Union (IoU). If the IoU exceeds a predefined threshold $\tau$, we adopt the intersection as the final prediction.
\begin{equation}
m_{i+n} = 
\begin{cases}
\texttt{IOU}(\widehat{m}_{i+n}^{\rightarrow}, \widehat{m}_{i+n}^{\leftarrow}), & \text{if } \texttt{IOU}(\cdot) \geq \tau \\
\widehat{m}_{i+n}^{\rightarrow} \cup \widehat{m}_{i+n}^{\leftarrow}, & \text{otherwise}
\end{cases}
\end{equation}
The OR operation in the second case accounts for cases with significant motion transitions, where the left and right predictions might deviate spatially or structurally. In such a case, two masks propagated from the left and the right considerably differ in either spatial location or shape. 

In rare cases, SAM2 may fail to propagate a mask from one direction, resulting in an empty prediction. We handle this with a fallback mechanism. Specifically, if the mask being propagated from the left is empty, we assign the right mask to the prediction of the current frame. Otherwise, we assign the left mask. Mathematically, it can be written as,
\begin{equation}
m_{i+n} =
\begin{cases}
\widehat{m}_{i+n}^{\leftarrow}, & \text{if } \widehat{m}_{i+n}^{\rightarrow} = \varnothing \\
\widehat{m}_{i+n}^{\rightarrow}, & \text{otherwise}
\end{cases}
\end{equation}
This bidirectional temporal fusion ensures temporally consistent masks with reduced compute overhead. It not only aligns with existing segmentation-based evaluation protocols but also introduces robustness in dynamic scenes by leveraging context from both past and future observations.

%% file: sections/4-dataset.tex
\begin{figure}[!t]
    \centering
    \includegraphics[width=1.0\textwidth]{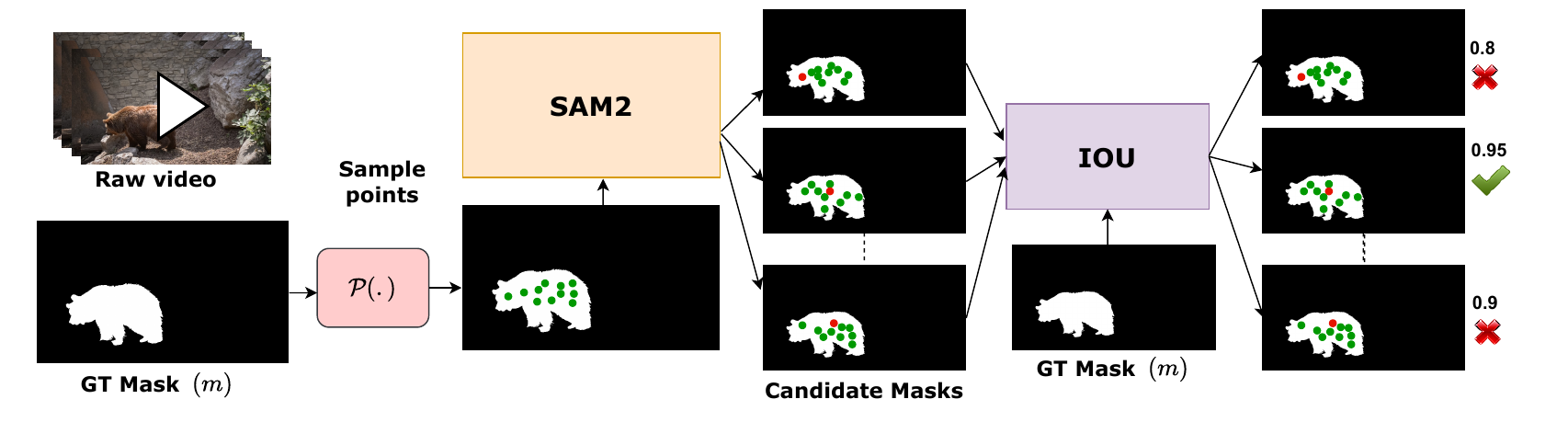}
    \caption{\textbf{\method annotation pipeline: } We construct point-level supervision from frame-level masks using a semi-automatic process. For each frame, $k$ points are sampled on the mask and passed to SAM2 to generate candidate masks. The point with the highest-IoU candidate mask (w.r.t. ground truth) is selected as the optimal annotation.}

\label{fig:annotation-pipeline}
\end{figure}
\section{\method dataset}
\vspace{-0.5em}
\textbf{Training Data:} Our dataset is essential for training the model’s spatio-temporal pointing capabilities. It features 72k Video-caption pairs along with 100k object points. Our benchmark video dataset comes from different sources: Refer-YTVOS \cite{seo2020urvos}, Refer-DAVIS \cite{Perazzi2016}, MeViS \cite{MeViS}, GroOT \cite{nguyen2023type}, LaSOT \cite{fan2019lasot}, ViCaS-LGVIS \cite{athar2024vicas}, and Reason-VOS \cite{yan2024visa}.To create fine-grained data grounded in point coordinates, we develop a semi-automated annotation pipeline (see Fig.~\ref{fig:annotation-pipeline}) that ensures both high-quality and scalable annotations. Each of the above mentioned datasets features video-mask-expression triplets $(V, M, T)$ such that $V \in \mathbb{R}^{|\mathcal{T}|\times H \times W \times C}, M \in \{0, 1\}^{|\mathcal{T}| \times |\mathcal{O}| \times H \times W}$ with $|\mathcal{O}|$ denoting the number of unique annotated objects in the frame $\mathcal{T}_i$. For each object $\mathcal{O}_{j} \in \mathcal{O}$ in frame $\mathcal{T}_{i} \in \mathbb{R}^{H \times W \times C}$, with a binary mask $m_{j} \in \{0,1\}^{H \times W}$, the goal is to extract a single highly representative point coordinate for the object. We sample $k$ candidate points within the mask, assigning each point $(x, y)$ a sampling probability proportional to its Euclidean distance to the nearest boundary pixel of the mask $m_j$, i.e.,
\begin{equation}
    P(x, y) \propto \min_{(x', y') \in \partial m_j} \; \left\| (x, y) - (x', y') \right\|_2
\end{equation}
where $\partial m_j$ denotes the set of boundary pixels of the mask. For each sampled point, we use SAM2 to predict a segmentation mask. We then compute the Intersection-over-Union (IoU) between each predicted mask and the corresponding ground truth mask $m_j$. The point coordinate whose predicted mask achieves the highest IoU is selected as the representative ground truth point for the object:
\begin{equation}
    p^* = \arg\max_{(x, y)} \; \mathrm{IoU} \left( \mathrm{SAM2}(x, y), \; m_j \right),
\end{equation}
where $\mathrm{SAM2}(x, y)$ denotes the predicted mask obtained using point $(x, y)$ as a prompt to SAM2.

\textbf{VPoS-Bench:}  To evaluate the generalization capabilities of \method, we introduce Video Pointing and Segmentation (VPoS-Bench), a curated benchmark test set comprising of $100$ video-caption pairs and $1k$ manually annotated object points.
For mask-based evaluations, we employ the SAM2 model to convert these point annotations into segmentation masks. The test benchmark encompasses diverse real-world scenarios, sourced from both open datasets \cite{caesar2020nuscenes,lin2024videogui,grauman2022ego4d,singh2024malmm,celltracking} and internal collections, spanning five categories: Cell Tracking, Egocentric Videos, Autonomous Driving, Video-GUI, and Robotics. Our benchmark also consists of a dedicated counting task sourced from \cite{kay2017kinetics} dataset. For additional details about the benchmark, refer to Sec. \ref{suppsec:Benchmark Details} of the Appendix.

%% file: sections/5-experiments.tex
\begin{figure}[!t]
    \centering
    \includegraphics[width=0.9\textwidth]{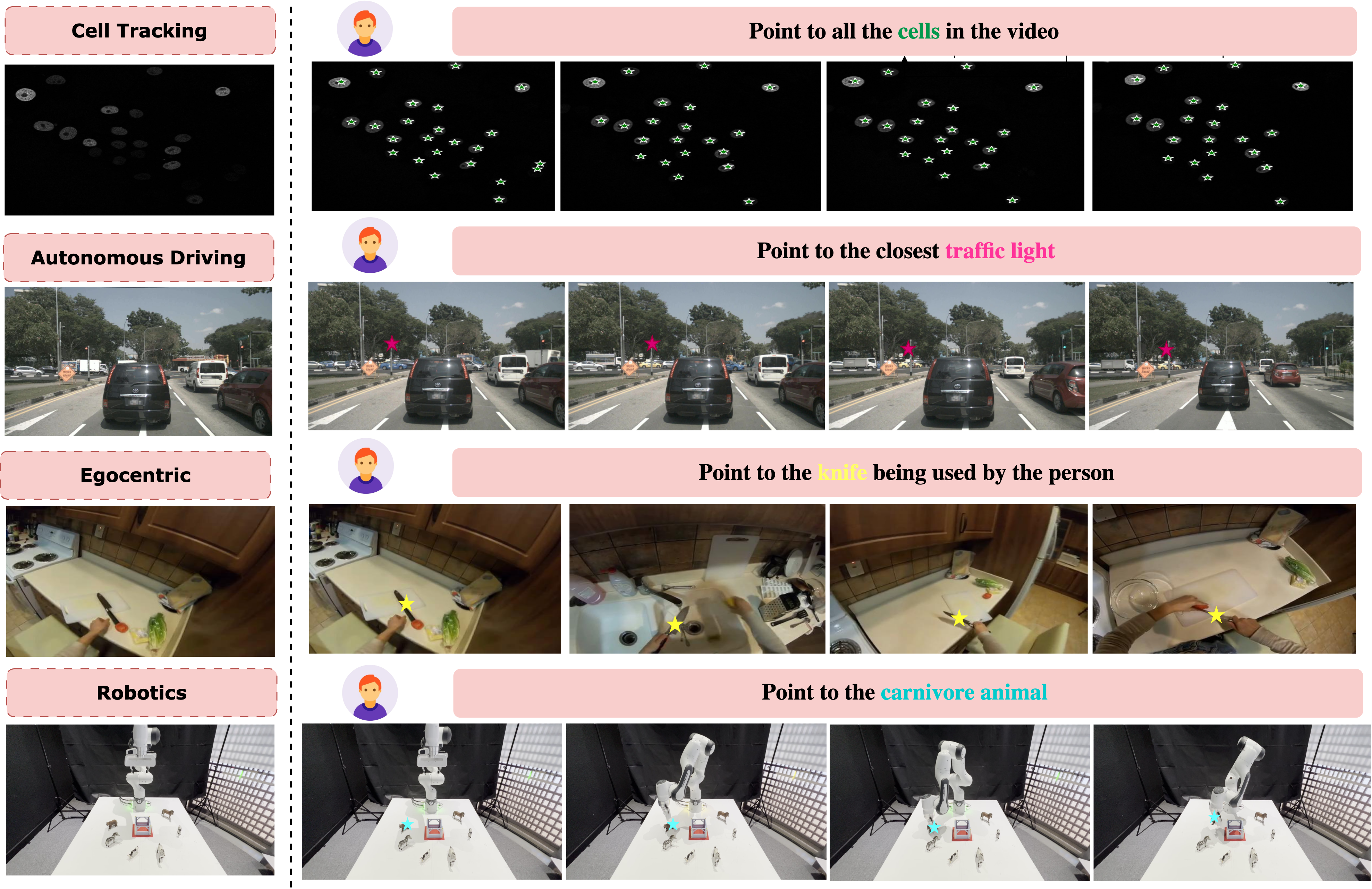}
    \caption{\method demonstrates robust generalization and fine-grained spatio-temporal grounding across diverse out-of-distribution scenarios from our proposed benchmark, for instance, correctly pointing to traffic lights (2\textsuperscript{nd} row) in challenging driving scenes despite never encountering such scenarios during training. (Please refer to Appendix~\ref{suppsec:Additionally qualitative results} for additional qualitative results.)}
\label{fig:benchmark-qualitative}
\vspace{-1.5em}
\end{figure}

\section{Experiments}
\label{Experimentation}
\vspace{-0.5em}
\noindent
\textbf{Implementation details. }
 \method follows the architecture of Molmo \cite{deitke2024molmo}. For the image encoder, we use a pretrained CLIP ViT-L/14 (336 × 336) \cite{radford2021learning} model. Our proposed Temporal Module is initialized from scratch. Our choice of LLM is the pretrained Qwen2 7B \cite{yang2024qwen2}. We train the model on $8$ NVIDIA A100 80GB GPUs. Learning rate of $1\text{e}^{-5}$ is used for the LLM, and $5\text{e}^{-6}$ for the vision encoder, visual projector and temporal module. We adopt a batch size of $1$ with $256$ gradient accumulation steps, and use AdamW optimizer \cite{loshchilov2019decoupled} following the fine-tuning recipe from \cite{deitke2024molmo}.\\
\noindent
\textbf{Tasks. } We evaluate \method on four challenging tasks: point grounding, counting, referring segmentation, and reasoning video object segmentation. For point grounding, we report performance on our proposed VPoS-Bench. 
For the counting task, we utilize videos from the Kinetics dataset \cite{kay2017kinetics}, where object counts range from $2 - 13$. For referring video segmentation, we use MeViS validation set \cite{MeViS} 
Refer-DAVIS-17 \cite{khoreva2019video} and Refer-Youtube-VOS \cite{Ref_ytvos} datasets. Finally, for reasoning segmentation, we evaluate our model on the ReasonVOS dataset \cite{bai2024one}.\\
\noindent
\textbf{Evaluation metrics. } For the Point Grounding task, we follow the evaluation protocol of Molmo \cite{deitke2024molmo} and report Precision, Recall, and F1 score.
For mask-based evaluations, we use Region Jaccard ($\mathcal{J}$), Boundary F-measure ($\mathcal{F}$), and their average ($\mathcal{J\&F}$). For the Counting task, we report Exact Matching Accuracy (EMA) and Mean Absolute Error (MAE) \cite{deitke2024molmo, DBLP:journals/prl/ZhaoWWLR23}.\\ 
\noindent
\textbf{Baselines.}
For point grounding, we compare \method with three strong baselines: VideoLISA\cite{bai2024one}, VideoGLaMM \cite{munasinghe2024videoglamm}, and Molmo+SAM2. To adapt Molmo for videos, we augment it with SAM2. For referring segmentation, we evaluate against VideoLISA, VideoGLaMM, and prior baselines. For counting, we compare \method with both closed-source (GPT-4o \cite{OpenAI2023ChatGPT}) and open-source models \cite{deitke2024molmo, bai2025qwen2,gemmateam2024gemma}).
For further experimentation details please refer to Appendix \ref{suppsec:Additional Experimental Details}.
\subsection{Main experimentation results}
\label{sec:main_results}
\noindent 
\textbf{Point Grounding. }
The point grounding task focuses on accurately identifying the spatial coordinates of a queried object within video frames. As depicted in Fig.~\ref{fig:pointing_metrics}, \method demonstrates superior performance in localizing target points, as evidenced by its significantly higher Precision, Recall, and F1 scores compared to Molmo. This performance gap can be attributed to Molmo’s training on static frames, which limits its ability to handle temporal variations. In dynamic video inputs, where object presence and position may vary across frames, Molmo struggles, whereas \method effectively addresses this challenge by leveraging temporal context. Furthermore, \method outperforms all baseline models across each subtask from our benchmark, as evident from higher $\mathcal{J}$, $\mathcal{F}$, and the combined $\mathcal{J\&F}$ metric (Table \ref{table:videomolmo-benchmark}). Qualitative results in Fig.\ref{fig:benchmark-qualitative} further validate the robustness of \method, showcasing its ability to accurately localize objects across diverse and out-of-distribution scenarios.

\textbf{Object Counting. }
Our benchmark introduces a dedicated object counting task, a capability essential to many real-world video understanding applications. The number of objects in our benchmark ranges from 2 to 13, therefore demanding enhanced temporal and spatial understanding. For this evaluation, we compare \method against both open-source and proprietary models, as shown in Table~\ref{tab:counting-benchmark}.
\method achieves state-of-the-art performance, significantly outperforming all baselines in terms of both Mean Absolute Error (MAE) and Exact Match Accuracy (EMA). Notably, \method also surpasses advanced proprietary model such as GPT-4o, underscoring its strength as a specialized counting model. We attribute this success to the explicit training of VideoMolmo on our proposed training dataset, which includes diverse scenarios with 0 to 10 objects per video, enabling the model to develop a fine-grained understanding of multi-object scenes.
\vspace{0.5em}

\input{tables/ref_davis}
\input{tables/reason_vos}

\input{tables/benchmark}
\input{tables/pointing}

\noindent 
\textbf{Referring Segmentation.} 
For referring video segmentation, the goal is to localize specific object instances in a video based on a given phrase. Table~\ref{table:ref-seg-merged} presents results across three standard datasets. On the MeViS benchmark, which involves motion-guided segmentation with multiple objects, \method outperforms all baselines by a notable margin, demonstrating its effectiveness in grounding complex, multi-object scenes. This advantage stems in part from the simplicity and efficiency of \method's point-based supervision as reflected in its superior $\mathcal{J}$, $\mathcal{F}$, and $\mathcal{J\&F}$ scores, which contrasts with recent methods like VideoGLaMM \cite{munasinghe2024videoglamm} and VideoMolmo \cite{bai2024one} that rely on dense, pixel-level mask prediction where precise delineation between objects becomes challenging (Fig. \ref{fig:teaser}).
\method also achieves superior performance on Refer-DAVIS-17 and Refer-Youtube-VOS. Notably, VideoGLaMM performs competitively on Refer-Youtube-VOS which features fast-moving objects, benefiting from its dual encoder architecture that integrates spatial and temporal features. However, \method, despite using a single encoder with point-based supervision, surpasses these strong baselines which could be attributed to its temporal module capturing past inference, novel post-processing technique, and point grounding training paradigm.

\noindent 
\textbf{Reasoning Video Object Segmentation. }
Table~\ref{table:reason-vos} presents a comparative analysis of \method against existing approaches on the ReasonVOS benchmark, which emphasizes complex reasoning, temporal comprehension, and consistency across frames, making it particularly challenging. Prior methods perform noticeably worse, largely due to their limited temporal and fine-grained reasoning capabilities. While VideoLISA incorporates spatio-temporal cues, it still falls short of \method. This performance gap highlights \method's architectural strengths, specifically its dedicated temporal module providing rich spatio-temporal contextual understanding. 
\input{tables/temporal}
\vspace{-0.5em}
\subsection{Ablations and Analysis}
\noindent
\textbf{Effect of Temporal Module.} We conduct an ablation study to evaluate the effectiveness of different temporal module variants on the Refer-DAVIS benchmark (Table~\ref{table:ablation-temporal-module}). Using a single frame or simple feature fusion methods such as addition or token-space concatenation yields relatively lower performance compared to our proposed cross-attention-based temporal module as it enables dynamic and selective integration of relevant features across frames, allowing the model to focus on temporally coherent and semantically meaningful cues critical for accurate grounding.

\noindent
\textbf{Ablation on Temporal Mask Fusion:}
To enable efficient and temporally consistent segmentation, we evaluate various strategies for combining masks propagated from the sampled frames. As shown in Table~\ref{table:ablation-mask-strategies}, naive strategies like preferring left/right predictions or computing mask intersections result in suboptimal performance, either due to loss of temporal context or overly conservative fusion. Our proposed bidirectional fusion strategy outperforms all baselines by adaptively reconciling forward and backward propagated masks based on their agreement (IoU). Our fallback mechanism ensures robustness against failure cases where one of the propagated masks is missing. This approach achieves a significant improvement in $\mathcal{J\&F}$ score (72.45), demonstrating its effectiveness. 

\noindent
\textbf{Effect of context-length in temporal module:}
To analyze the effect of context length in the temporal module, we evaluate \method on the Refer-DAVIS benchmark (Fig. \ref{fig:temporal-module}). We observe that there is a consistent increase in $J \& F$ as the context length increases from $1 \to 4$, indicating that incorporating more temporal information enhances the model's spatio-temporal reasoning. However, there is a slight drop in accuracy at $l = 5$, suggesting that adding more frames may introduce redundancy or noise rather than useful context.
Please refer to Appendix \ref{suppsec:Additional Ablation Results} for additional ablations.


%% file: tables/ref_davis.tex
\begin{table}[t]
\centering
\setlength{\tabcolsep}{7pt}
\caption{Performance comparison on Refer-DAVIS-17, Refer-Youtube-VOS, and MeViS benchmarks. VideoMolmo consistently improves referring video object segmentation across datasets.}
\resizebox{0.9\textwidth}{!}{
\begin{tabular}{l|ccc|ccc|ccc}
    \toprule
    \textbf{Model} 
    & \multicolumn{3}{c|}{\textbf{Refer-DAVIS-17}} 
    & \multicolumn{3}{c|}{\textbf{Refer-Youtube-VOS}} 
    & \multicolumn{3}{c}{\textbf{MeViS}} \\
    \cmidrule(lr){2-4} \cmidrule(lr){5-7} \cmidrule(lr){8-10}
    & $\mathcal{J}$ & $\mathcal{F}$ & $\mathcal{J\&F}$ 
    & $\mathcal{J}$ & $\mathcal{F}$ & $\mathcal{J\&F}$ 
    & $\mathcal{J}$ & $\mathcal{F}$ & $\mathcal{J\&F}$ \\
    \midrule
    LISA-7B \cite{lai2023lisa}         & 61.9 & 54.9 & 58.4 & 50.6 & 49.7 & 50.2 & --   & --   & --   \\
    LISA-13B \cite{lai2023lisa}        & 64.6 & 56.8 & 60.7 & 53.0 & 52.1 & 52.6 & --   & --   & --   \\
    TrackGPT-7B \cite{zhu2023tracking} & 67.0 & 59.4 & 63.2 & 57.4 & 55.3 & 56.4 & --   & --   & --   \\
    TrackGPT-13B \cite{zhu2023tracking}& 70.4 & 62.7 & 66.5 & 60.8 & 58.1 & 59.5 & --   & --   & --   \\
    VideoLISA \cite{bai2024one}        & 72.7 & 64.9 & 68.8 & 65.7 & 61.7 & 63.7 & 41.3 & 47.6 & 44.4 \\
    VideoGLaMM \cite{munasinghe2024videoglamm} & \textbf{73.3} & 65.6 & {69.5} & 65.4 & 68.2 & 66.8 & 42.1 & 48.2 & 45.2 \\
    Molmo \cite{deitke2024molmo}+SAM2 \cite{ravi2024sam} & 65.3 & 72.2 & 68.8 & 61.0 & 66.2 & 63.6 & 44.4 & 49.4 & 46.9 \\
    \midrule
    \rowcolor{violet!10} 
    VideoMolmo                    & 71.3 & \textbf{73.6} & \textbf{72.5} 
                                       & \textbf{65.6} & \textbf{69.1} & \textbf{67.3} 
                                       & \textbf{51.2} & \textbf{56.6} & \textbf{53.9} \\
    \bottomrule
\end{tabular}
}
\label{table:ref-seg-merged}
\vspace{-1em}
\end{table}

%% file: tables/reason_vos.tex
\begin{table}[t]
\begin{minipage}[t]{0.48\textwidth}
    \centering
    \caption{Performance comparison of VideoMolmo on the ReasonVOS benchmark.}
    \resizebox{\textwidth}{!}{
    \begin{tabular}{lccc}
    \toprule
    \textbf{Model} & $\mathcal{J}$ & $\mathcal{F}$ & $\mathcal{J\&F}$ \\ 
    \midrule
    LISA \cite{lai2023lisa} & 33.1 & 29.1  &  31.1 \\
    VideoLISA \cite{bai2024one} & 49.9 & 45.1 & 47.5 \\
    VideoGLaMM \cite{munasinghe2024videoglamm} & 40.5 & 27.2 & 33.9 \\
    Molmo \cite{deitke2024molmo} + SAM2 \cite{ravi2024sam} & 43.5 & 47.8 & 45.7 \\
    \hline
    \rowcolor{violet!10}
    VideoMolmo & \textbf{48.7} & \textbf{53.4} & \textbf{51.1} \\
    \bottomrule
    \end{tabular}
    }
    \label{table:reason-vos}
\end{minipage}
\hfill
\begin{minipage}[t]{0.45\linewidth}
    \centering \renewcommand{\arraystretch}{1.1}
    \caption{Performance comparison of VideoMolmo on the counting benchmark.}
    \resizebox{0.85\textwidth}{!}{
    \begin{tabular}{lc@{\hskip 12pt}c}
    \toprule
    \textbf{Model} & \textbf{MAE $\downarrow$} & \textbf{EMA $\uparrow$} \\
    \midrule
    GPT-4o \cite{openai2024gpt4o}       & 0.76 & 60.0 \\
    Gemma3-12B \cite{gemmateam2024gemma}    & 0.96 & 43.3 \\
    Qwen2.5-VL-7B \cite{bai2025qwen2} & 0.83 & 50.0 \\
    Molmo ~\cite{deitke2024molmo}        & 1.21 & 49.3 \\
    \rowcolor{violet!10}\hline
    VideoMolmo    & \textbf{0.72} & \textbf{73.3} \\
    \bottomrule
    \end{tabular}
    }\renewcommand{\arraystretch}{1.0}
    \label{tab:counting-benchmark}
\end{minipage}
\vspace{-1em}
\end{table}

%% file: tables/benchmark.tex
\begin{table}[!t]
\centering
\caption{Performance of various models on five subtasks of VPoS-Bench (Ego4D, Robotics, Autonomous, Cells, VideoGUI)}
\LARGE 
\setlength{\tabcolsep}{8pt}
\resizebox{\textwidth}{!}{ 
\begin{tabular}{l|ccc|ccc|ccc|ccc|ccc}
\toprule
\textbf{Model} 
& \multicolumn{3}{c|}{\textbf{Ego4D}} 
& \multicolumn{3}{c|}{\textbf{Robotics}} 
& \multicolumn{3}{c|}{\textbf{Autonomous}} 
& \multicolumn{3}{c|}{\textbf{Cells}} 
& \multicolumn{3}{c}{\textbf{VideoGUI}} \\
& $\mathcal{J}$ & $\mathcal{F}$ & $\mathcal{J\&F}$
& $\mathcal{J}$ & $\mathcal{F}$ & $\mathcal{J\&F}$
& $\mathcal{J}$ & $\mathcal{F}$ & $\mathcal{J\&F}$
& $\mathcal{J}$ & $\mathcal{F}$ & $\mathcal{J\&F}$
& $\mathcal{J}$ & $\mathcal{F}$ & $\mathcal{J\&F}$ \\
\midrule
VideoLISA~\cite{bai2024one}        & 47.1 & 41.1 & 44.2 & 3.9 & 2.0 & 2.9 & 34.7 & 22.0 & 28.4 & 18.9 & 3.3 & 11.1 & 65.3 & 39.4 & 52.4 \\
VideoGLaMM~\cite{munasinghe2024videoglamm}                        & 47.2 & 40.4 & 43.8 & 15.3 & 10.3 & 12.8 & 31.4 & 17.7 & 24.8 & 11.8 & 7.8 & 9.8 & 58.7 & 32.5 & 45.6 \\
Molmo~\cite{deitke2024molmo}   +SAM2~\cite{ravi2024sam}                         & 50.6 & 50.1 & 50.4 & 27.8 & 25.6 & 26.7 & 49.5 & 47.5 & 48.5 & 20.8 & 7.6  & 14.2 & 60.2 & 55.9 & 58.0 \\ \hline
\rowcolor{violet!10} VideoMolmo   & \textbf{55.5} & \textbf{54.3} & \textbf{54.9} & \textbf{29.1} & \textbf{26.2} & \textbf{27.6}
& \textbf{57.1} & \textbf{57.7} & \textbf{57.4} &
\textbf{25.4} & \textbf{13.1} & \textbf{19.2} & \textbf{68.9} & \textbf{62.4} & \textbf{65.7} \\
\bottomrule
\end{tabular}
}
\label{table:videomolmo-benchmark}
\vspace{0em}
\end{table}

%% file: tables/pointing.tex
\begin{figure}[!t]
    \centering
    \begin{minipage}[t]{0.45\textwidth}
        \centering
        \includegraphics[width=0.95\textwidth]{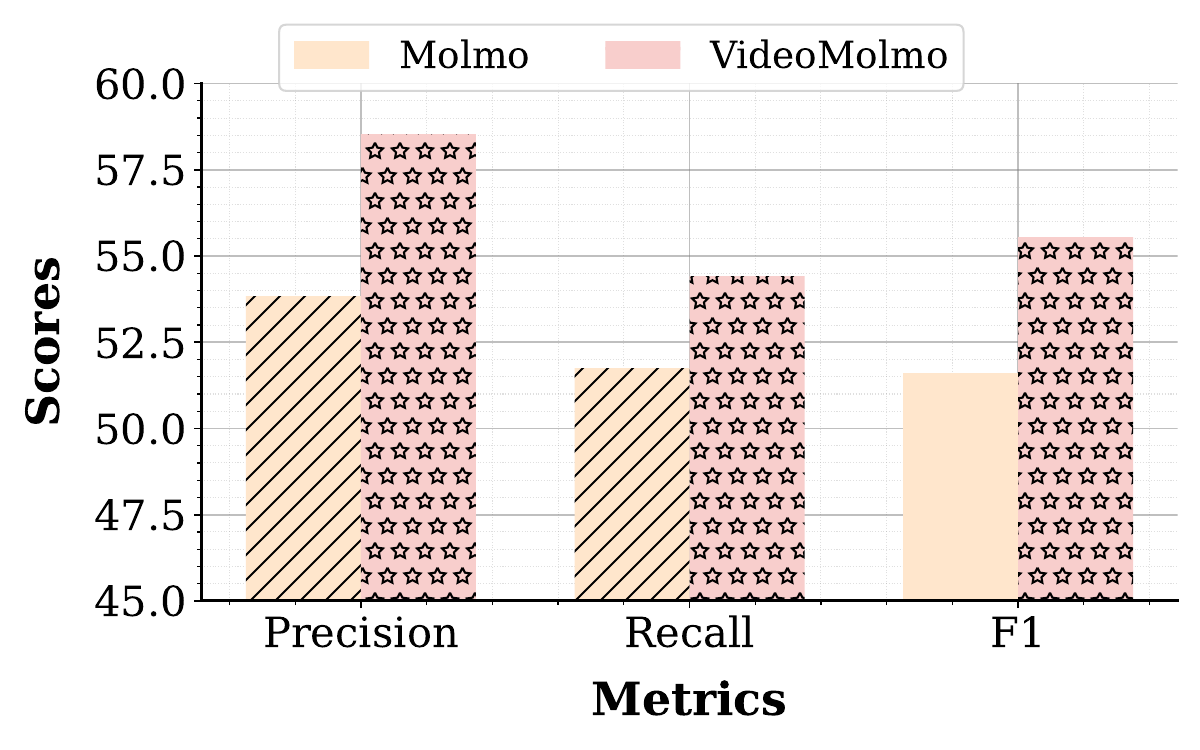}
         \vspace{-0.5em}
        \caption{Performance comparison of VideoMolmo on point grounding.}
        \label{fig:pointing_metrics}
    \end{minipage}
    \hfill
    \begin{minipage}[t]{0.45\textwidth}
        \centering
        \includegraphics[width=0.9\textwidth]{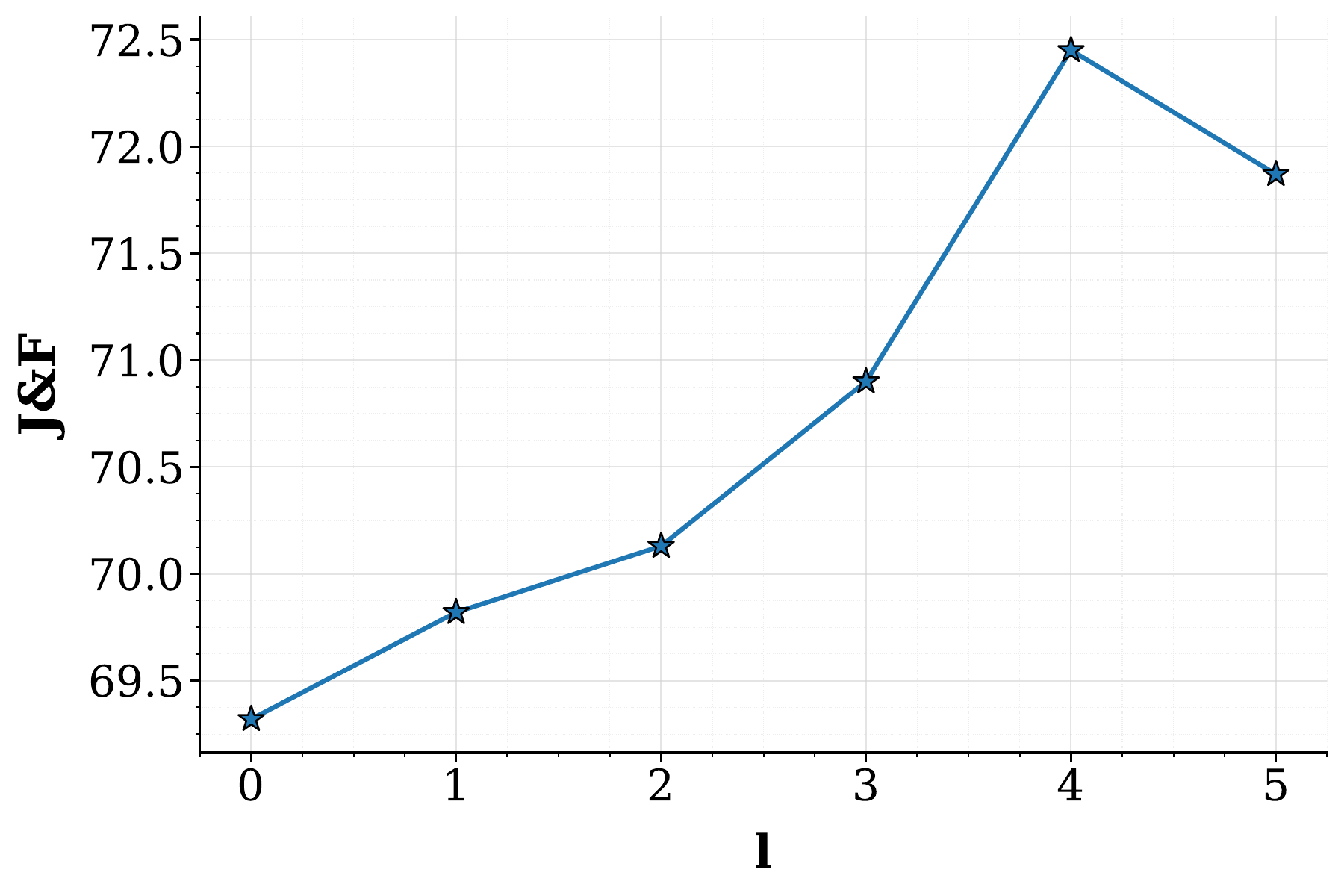}
        \vspace{-0.5em}
        \caption{Effect of temporal module context length on segmentation accuracy in Refer-DAVIS benchmark.}
        \label{fig:temporal-module}
    \end{minipage}
    \vspace{-2em}
\end{figure}

%% file: tables/temporal.tex
\begin{table}[!t]
\centering
\small
\setlength{\tabcolsep}{8pt}

\begin{minipage}{0.42\textwidth}
    \centering
    \caption{Effect of different temporal mask fusion strategies on Refer-DAVIS dataset.}
    \resizebox{\textwidth}{!}{
    \begin{tabular}{lccc}
    \toprule
    \textbf{Strategy} & $\mathcal{J}$ & $\mathcal{F}$ & $\mathcal{J\&F}$ \\ 
    \midrule
    Prefer left    & 67.31 & 73.38 & 70.34 \\
    Prefer right   & 67.05 & 71.69 & 69.37 \\
    Intersection   & 60.20 & 63.58 & 61.89 \\
    Larger mask    & 70.40 & 72.91 & 71.65 \\
    Smaller mask   & 64.10 & 67.18 & 65.64 \\
    \hline
    \rowcolor{violet!10}
    VideoMolmo & \textbf{71.27} & \textbf{73.63} & \textbf{72.45} \\
    \bottomrule
    \end{tabular}
    }
    \label{table:ablation-mask-strategies}
\end{minipage}
\hfill
\begin{minipage}{0.48\textwidth}
    \centering \renewcommand{\arraystretch}{1.15}
    \caption{Ablation on different variants of the temporal module on the Refer-DAVIS dataset.}
    \resizebox{\textwidth}{!}{
    \begin{tabular}{lccc}
    \toprule
    \textbf{Variant} & $\mathcal{J}$ & $\mathcal{F}$ & $\mathcal{J\&F}$ \\ 
    \midrule
    Single frame   & 66.04 & 72.60 & 69.32 \\
    Addition       & 66.13 & 72.97 & 69.55 \\
    Concatenation  & 65.34 & 72.06 & 68.71 \\
    \hline
    \rowcolor{violet!10}
    VideoMolmo & \textbf{71.27} & \textbf{73.63} & \textbf{72.45} \\
    \bottomrule
    \end{tabular}
    }\renewcommand{\arraystretch}{1.0}

    \label{table:ablation-temporal-module}
\end{minipage}
\vspace{-1.6em}
\end{table}

%% file: sections/7-conclusion.tex

\vspace{-0.5em}
\section{Conclusion}
\vspace{-0.5em}
We present \method, an LMM for fine-grained spatio-temporal pointing conditioned on textual queries. It leverages a temporal module that incorporates temporal cues from preceding frames and a novel bidirectional post-processing strategy for robust mask prediction. To enable training, we curate a large-scale dataset using a semi-automatic annotation pipeline. \method shows strong generalization and consistently outperforms state-of-the-art models across diverse and out-of-distribution tasks, including point grounding, object counting, referring segmentation, and reasoning segmentation.\\
\noindent
\textbf{Limitations and Future Work. }
\label{limitations}
\method demonstrates strong spatio-temporal grounding performance, excelling in fine-grained localization without explicit pixel-level mask supervision. However, its performance is relatively limited on videos with fast-moving objects, such as those in Refer-Youtube-VOS (Table~\ref{table:ref-seg-merged}), due to single-frame processing during training—an efficiency-driven design constrained by architectural and computational limits. Additionally, \method relies on SAM2, making mask quality dependent on SAM2’s performance from point predictions. Predicting a single point per object can also yield suboptimal masks (see~\ref{suppsec:Additionally qualitative results}). Future work may include joint multi-frame training with improved sampling strategies and extending \method to predict multiple points per object to enhance mask quality and further benefit downstream applications.

\section{Acknowledgement} 
The computations were enabled by resources provided by NAISS at Alvis partially funded by Swedish Research Council through grant agreement no. 2022-06725, LUMI hosted by CSC (Finland) and LUMI consortium, and by Berzelius resource provided by the Knut and Alice Wallenberg Foundation at the NSC.

%% file: sections/99-appendix.tex
\appendix
\setcounter{page}{1}
\section{Appendix}

\subsection{Additional Ablations }
\label{suppsec:Additional Ablation Results}

\subsubsection{Training Ablations}
\noindent



In addition to the ablation studies presented in the main paper, we conduct further investigations into the impact of language backbone choice, parameter tuning, and numerical precision on the Ref-DAVIS dataset, as summarized in Table~\ref{tab:videomolmo_variants}.

In the first row, we assess the effect of replacing the Qwen-7B language model with Olmo-7B. The resulting drop in $\mathcal{J}\&\mathcal{F}$ score highlights Qwen-7B’s superior grounding capabilities which is consistent with the observational findings reported in \cite{deitke2024molmo}. This emphasizes the importance of selecting an Qwen-7b LLM with strong multimodal alignment for visual grounding tasks.

\input{tables/videomolmo_variants}

Next, we investigate the impact of end-to-end training. In the second row, we freeze the video encoder and train only the LLM’s projection layers by integrating LoRA adapters. This lightweight training strategy significantly underperforms compared to the fully fine-tuned model (last row), validating our hypothesis that joint optimization of all components is essential for capturing the temporal nuances required for precise point grounding.

Finally, we examine the effect of training precision. In the third row, we use 16-bit floating point precision which is commonly adopted to save memory and accelerate training. However, we find that this leads to a notable degradation in performance. In contrast, training with full 32-bit precision (last row) enhances the model’s capacity to learn fine-grained spatial and temporal cues, consistent with prior observations in \cite{deitke2024molmo}.

Together, these ablations underline the significance of careful backbone selection, full end-to-end optimization, and high-precision training for achieving robust and fine-grained visual grounding in VideoMolmo.

\subsubsection{Inference Ablations}
\noindent
\textbf{Sampling rate $k$:} 
As described in the main paper, we adopt a frame sampling rate of $k = |\mathcal{T}|$ for the Molmo+SAM2 baseline during inference, which means that we take the first frame prediction and use SAM2 to propagate the mask across all the frames. This choice is motivated by performance on the Refer-DAVIS-17 dataset, where Molmo+SAM2 achieves its highest $\mathcal{J}\&\mathcal{F}$ score of $67.69$ at this value. However, our analysis in Table \ref{table:ablation-molmo-sam2-} reveals that the optimal sampling rate is not universal, it varies across datasets.
To ensure a fair and competitive comparison with our proposed VideoMolmo, we conduct additional ablations on the sampling rate $k$ for Molmo+SAM2 across the Refer-YouTube-VOS and MeViS datasets. We find that a lower sampling rate of $k=3$ yields the best performance on Refer-YouTube-VOS, while $k=1$ proves optimal on MeViS. Despite this tuning, VideoMolmo consistently outperforms Molmo+SAM2 under each dataset’s optimal configuration.
Interestingly, across all three datasets, we observe a consistent decline in performance as the sampling rate increases. This is particularly evident at $k = 30$, where the baseline performs starts dropping. These findings further highlight the robustness of VideoMolmo in leveraging temporal context, even when competing baselines are tuned to their best-performing configurations.

We further ablate the effect of sampling rate on our proposed VideoMolmo. While our main results on the Refer-YouTube-VOS benchmark in the main paper are reported using a sampling rate of $k=5$, we acknowledge that this choice, although consistent with the baseline configuration, may not be optimal for our method. As seen from the Table \ref{table:ablation-sampling_rate-ref-yt-vos}, VideoMolmo benefits from careful selection of sampling rate as we observe that a sampling rate $k=20$ yields the highest $\mathcal{J}\&\mathcal{F}$ score of $68.14$, compared to $67.33$ with $k=5$.

\input{tables/molmo_sampling_rate}

\textbf{Threshold $\tau$:}
Our proposed post-processing strategy, \textit{Bidirectional Temporal Mask Fusion}, enhances model performance by combining masks propagated from both right and left directions to achieve a robust temporal consensus. As described in Section~\ref{sec:post-processing} of the main paper, the fusion process is governed by a threshold hyperparameter $\tau$, which determines how agreement between the two masks is evaluated.
Specifically, when the Intersection-over-Union (IoU) between the forward and backward masks exceeds $\tau$, their intersection is used as the final mask, enforcing stricter agreement. Conversely, if the IoU falls below $\tau$, their union is taken, promoting flexibility in ambiguous regions. This mechanism balances precision and recall based on temporal consistency.
We ablate different values of $\tau$ in Table~\ref{table:ablation-threshold-ref-davis} to identify the most effective setting. The results indicate that $\tau = 0.7$ yields the best overall performance. However, the differences across values are relatively minor, underscoring the high quality and stability of the point predictions generated by VideoMolmo. This consistency highlights the robustness of our model in temporal point grounding, even under varying post-processing thresholds.
\input{tables/threshold_ref_ytvos}

\begin{figure}[!t]
    \centering
    \includegraphics[width=0.75\textwidth]{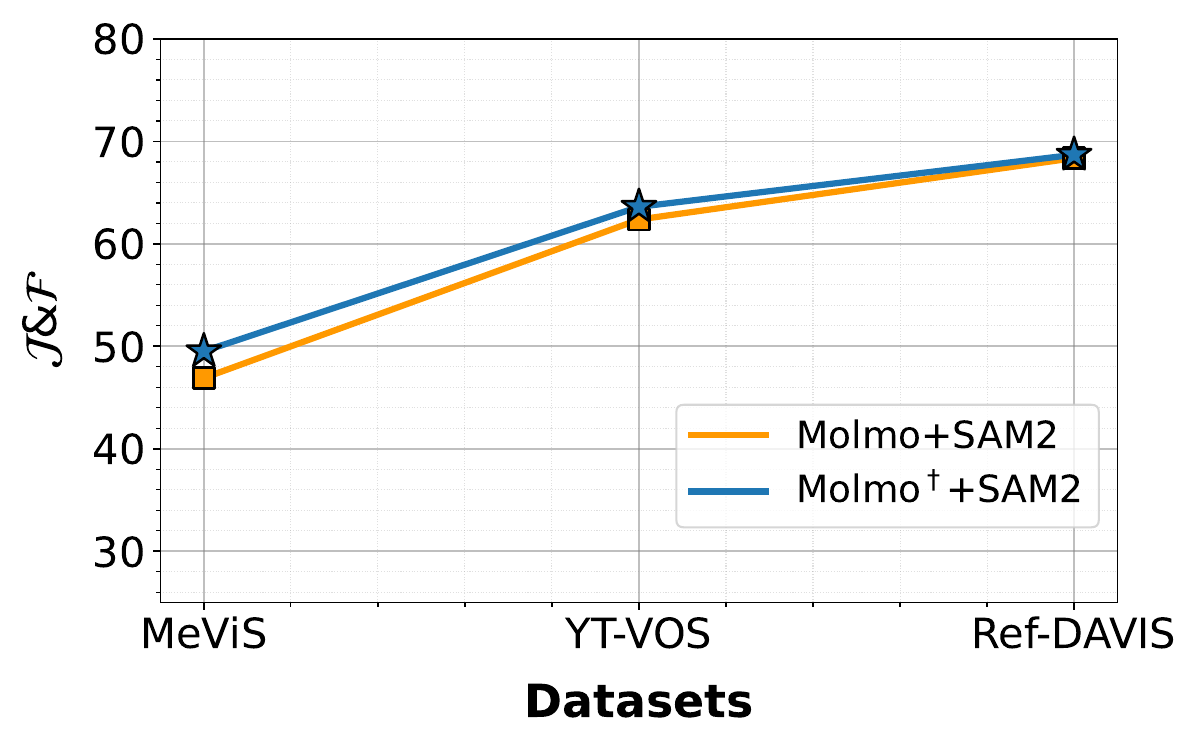}
    \caption{Effect of \textit{Bidirectional temporal mask fusion} on Molmo+SAM2 baseline.}
\label{fig:supp-molmo_vs_molmo_pp}
\end{figure}

\textbf{Effect of Post-processing on baselines: }
To assess the generalizability and effectiveness of our proposed \textit{Bidirectional Temporal Mask Fusion}, we integrate it with the Molmo+SAM2 baseline, resulting in an enhanced variant denoted as Molmo$^\dagger$+SAM2. As illustrated in Fig.~\ref{fig:supp-molmo_vs_molmo_pp}, this integration consistently improves performance across all three datasets in terms of $\mathcal{J\&F}$ score.
These results demonstrate that our post-processing strategy not only strengthens our own model but also benefits existing methods. The modular, plug-and-play nature makes it a valuable addition to any video grounding pipeline, improving temporal consistency and overall segmentation quality with minimal integration effort.

\subsection{Additional Experimentation Details }
\label{suppsec:Additional Experimental Details}
\method follows the architecture of Molmo \cite{deitke2024molmo}. For the image encoder, we use a pretrained CLIP ViT-L/14 (336 × 336) \cite{radford2021learning} model. We initialize the temporal module with Xavier Normalized weights for stable training.  Our choice of LLM is the pretrained Qwen2 7B \cite{yang2024qwen2}. We train the model on $8$ NVIDIA A100 80GB GPUs. Learning rate of $1\text{e}^{-5}$ is used for the LLM, and $5\text{e}^{-6}$ for the viion encoder, visual projector and temporal module. We adopt a batch size of $1$ with $256$ gradient accumulation steps, and use AdamW optimizer with $\beta = (0.9, 0.95)$ and   $\epsilon=1e^{-6}$, .
We train \method for $4$ epochs on $8$ NVIDIA A100 GPUs (80 GB each), consuming roughly 1,000 GPU-hours in total. Training runs in full 32-bit precision with a 10-step linear warmup, after which we follow a cosine learning-rate schedule; we also clip gradients to a maximum norm of $1.0$ to guard against unstable updates. For all inference and reported results, we use 4-bit precision.

\subsection{VPoS Bench}
\label{suppsec:Benchmark Details}
As mentioned in the main paper, we introduce Video Pointing and Segmentation (VPoS-Bench), a curated benchmark test set comprising of $100$ video-caption pairs and $1k$ manually annotated object points. To obtain mask annotations for evaluation, we use SAM2 to convert these point annotations into segmentation masks.
For mask-based evaluations, we employ the SAM2 model to convert these point annotations into segmentation masks. The test benchmark encompasses diverse real-world scenarios, sourced from both open datasets \cite{caesar2020nuscenes,lin2024videogui,grauman2022ego4d} and internal collections, spanning five categories: Cell Tracking, Egocentric Videos, Autonomous Driving, Video-GUI, and Robotics. Our benchmark also consists of a dedicated counting task sourced from \cite{kay2017kinetics} dataset. Below, we present details about each subset in VPoS-Bench.

\noindent
\textit{1) Cell Tracking: }Features internally sourced 12 microscopic videos with dynamic cellular structures, where precise localization of individual cells is essential for tasks like tracking cell division or counting. These videos are partially sourced from \cite{celltracking} and the remaining are requested internally.

\textit{2) Egocentric Videos: } Comprises 18 first-person videos capturing daily human-object interactions, enabling the assessment of grounded pointing in scenarios such as object manipulation and activity recognition. The egocentric videos in our test benchmark are derived from \cite{grauman2022ego4d} dataset.

\textit{3) Autonomous Driving: }Includes 13 urban driving scenes from nuScenes's dataset \cite{caesar2020nuscenes}, with complex environments, requiring accurate identification of specific road elements (e.g., traffic signals) to support navigation and safety systems.

\textit{4) Video-GUI: }Consists of 13 screen recordings from software applications, focusing on tasks like identifying and interacting with user interface elements based on textual instructions. The VideoGUI videos are sampled from VideoGUI dataset \cite{lin2024videogui}.

\textit{5) Robotics: } Encompasses 14 videos of robotic operations, emphasizing the need for precise object localization to execute commands such as "pick up the red block" or "press the top button." Few of the robotic videos in our benchmark are sourced from \cite{singh2024malmm}, and the remaining are sourced internally.


\subsection{Additional qualitative results}
\label{suppsec:Additionally qualitative results}
\begin{figure}[!h]
    \centering
    \includegraphics[width=\textwidth]{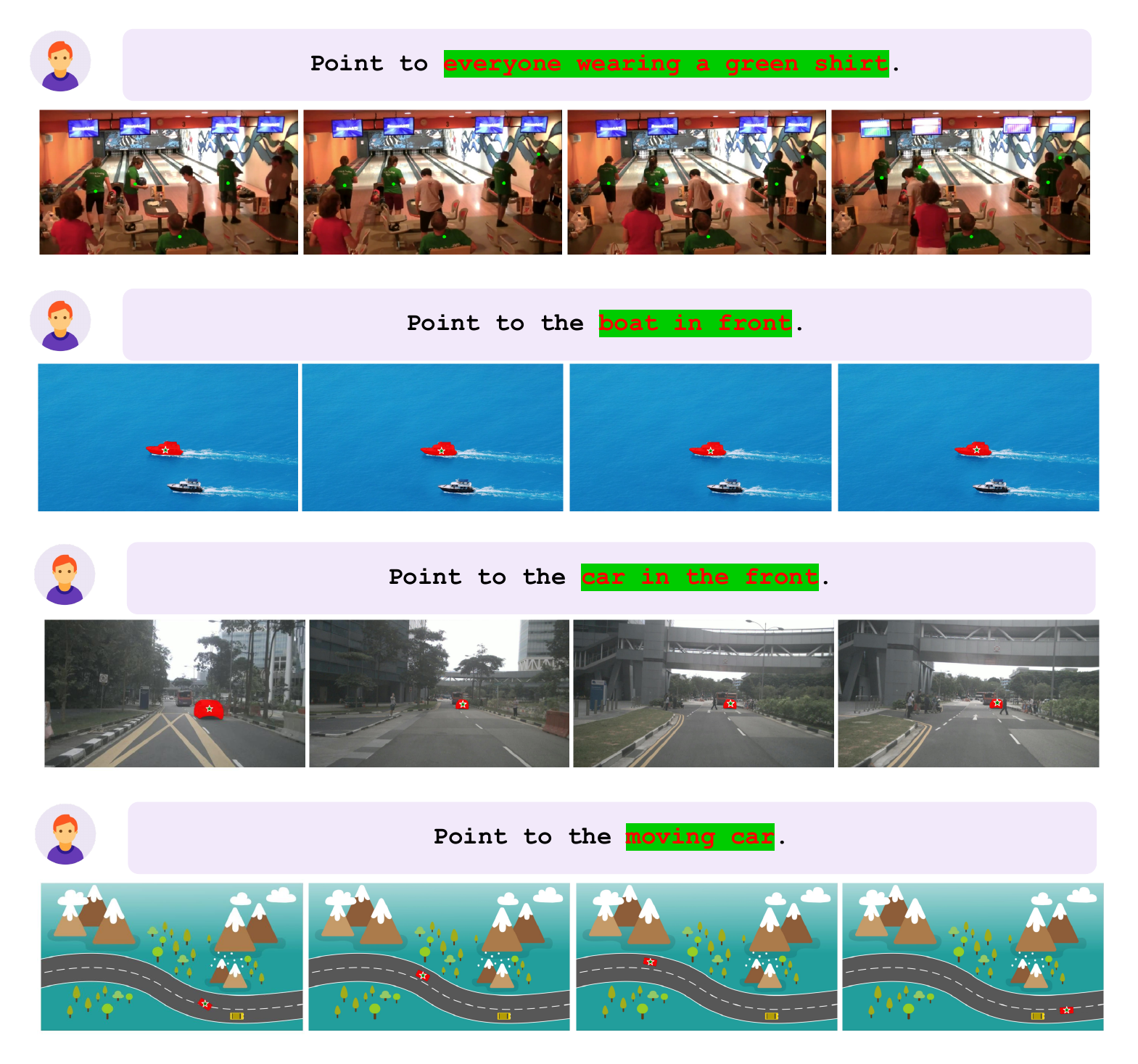}
    \caption{VPoS-Bench qualitative examples.}
\label{fig:supp-vpos-qualitative}
\end{figure}

\begin{figure}[!t]
    \centering
    \includegraphics[width=\textwidth]{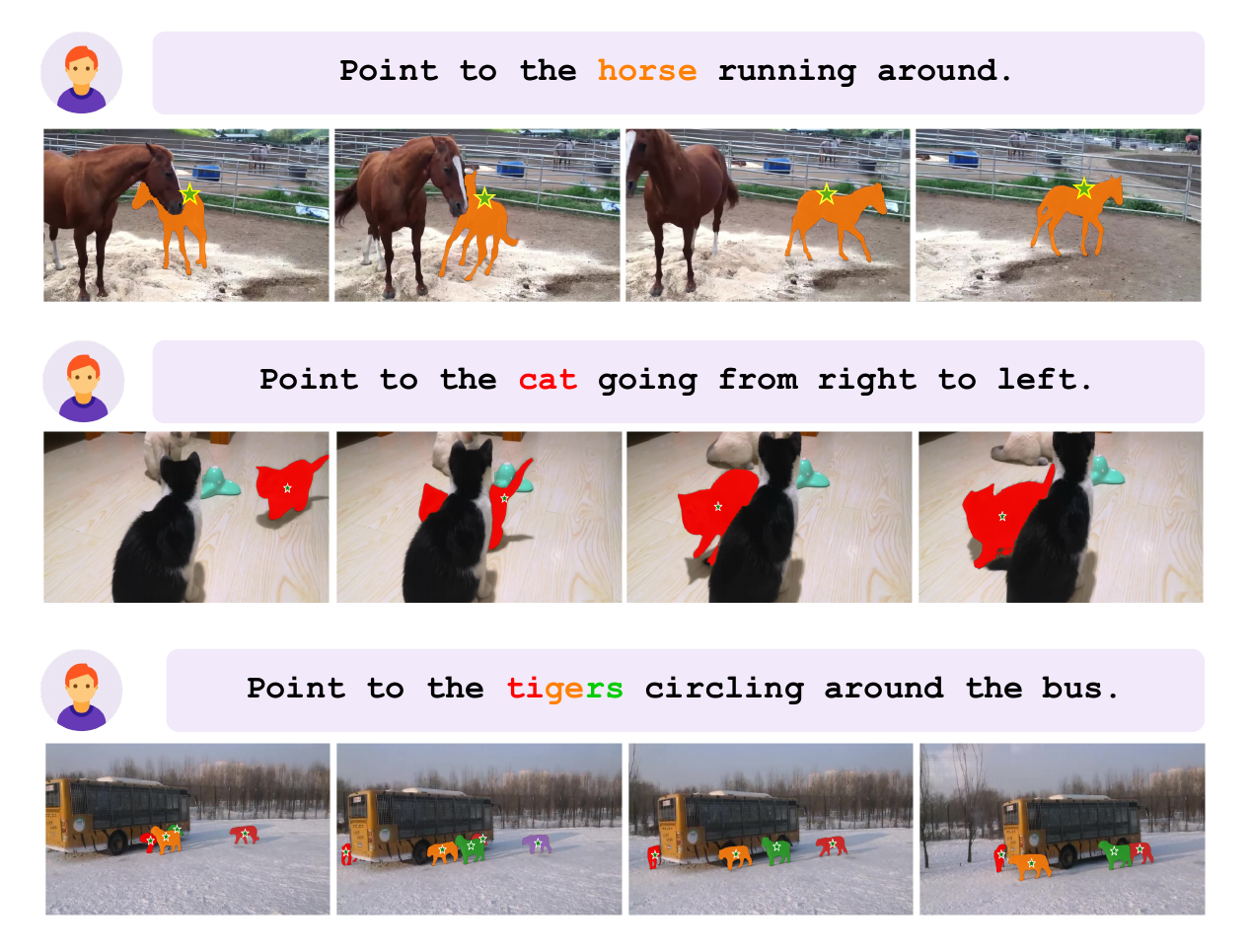}
    \caption{MeVis qualitative examples.}
\label{fig:supp-mevis-qualitative}
\end{figure}

\begin{figure}[!t]
    \centering
    \includegraphics[width=\textwidth]{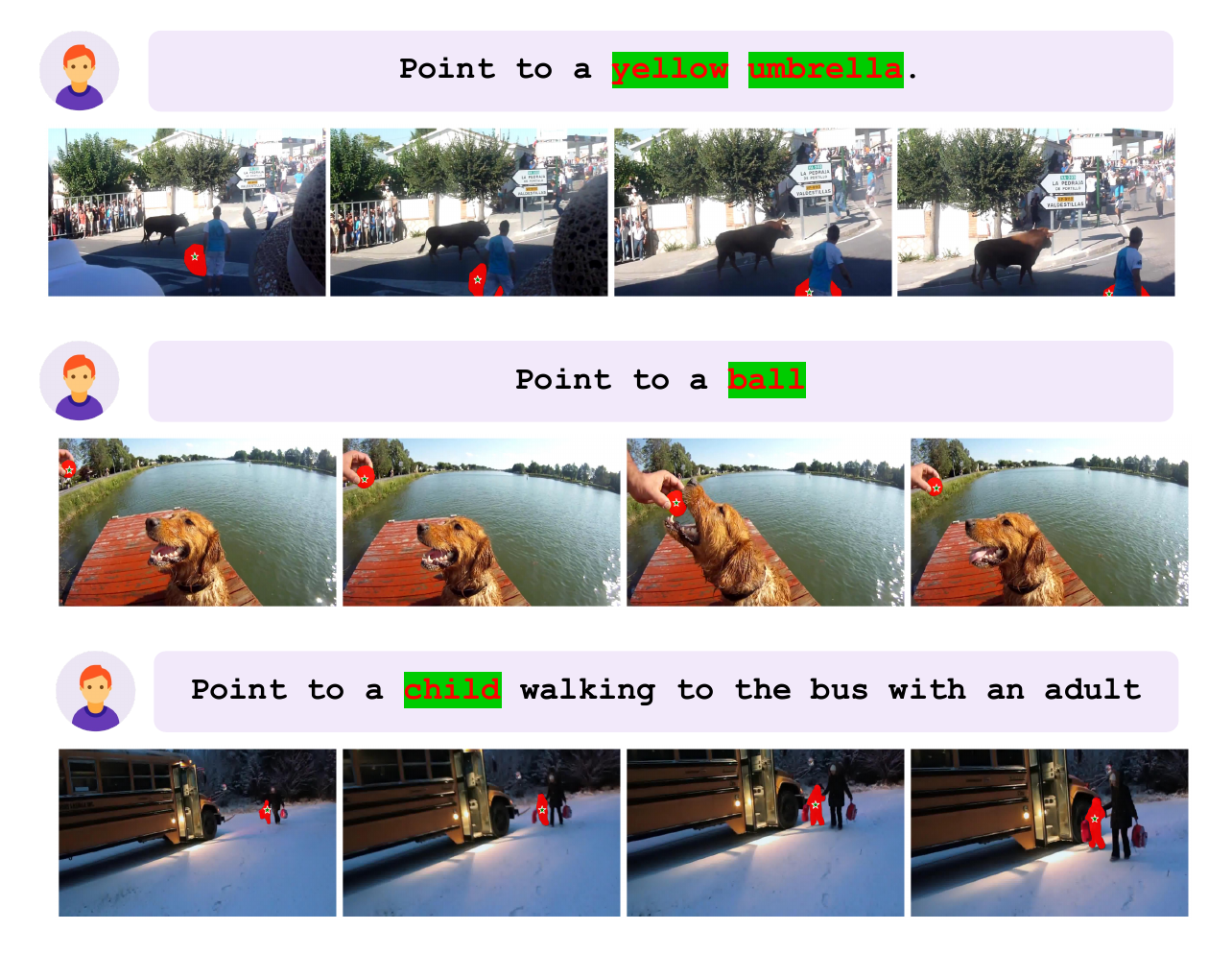}
    \caption{Refer-YouTube-VOS qualitative examples.}
\label{fig:supp-ytvos-qualitative}
\end{figure}

\begin{figure}[!t]
    \centering
    \includegraphics[width=\textwidth]{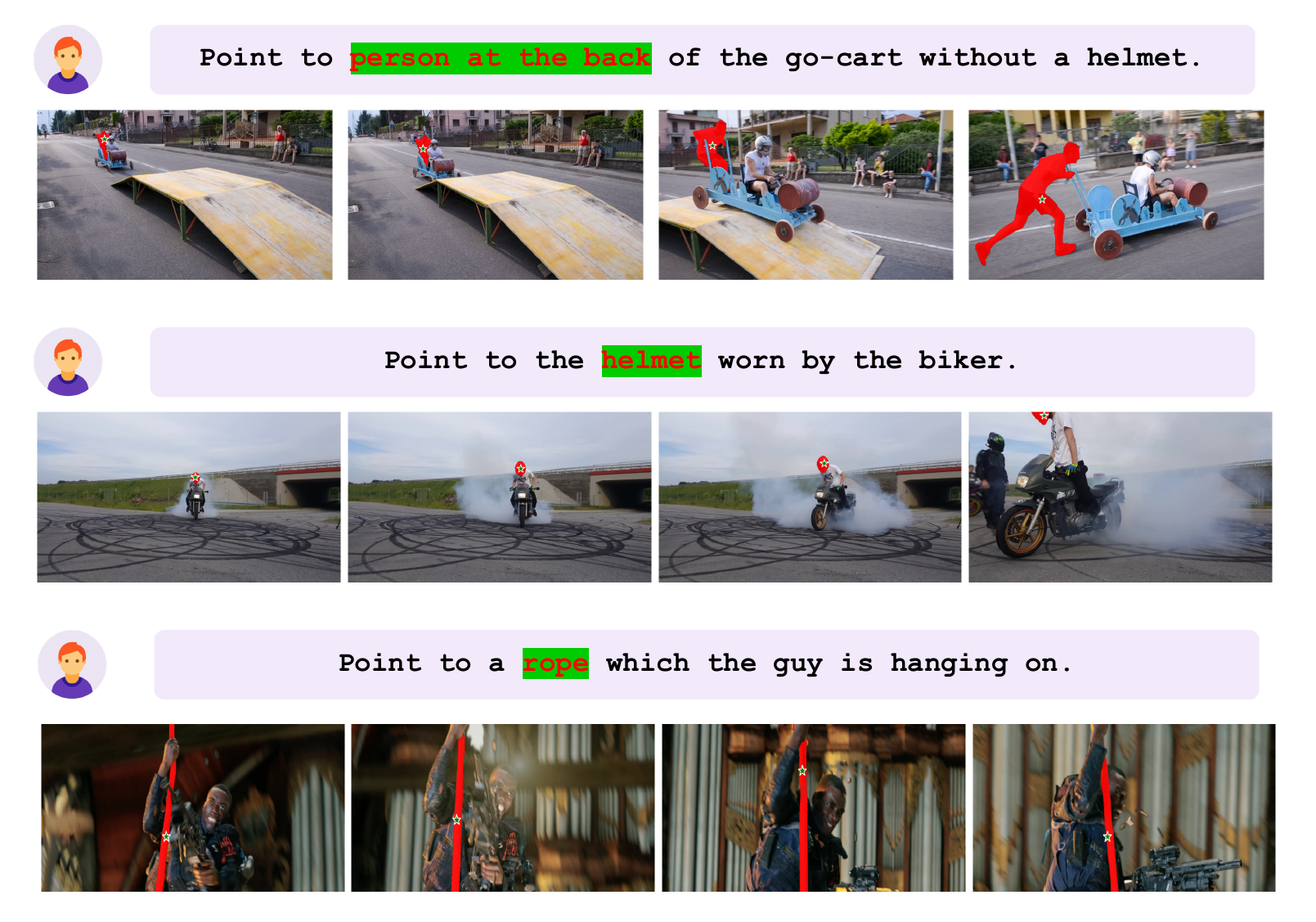}
    \caption{Refer-DAVIS qualitative examples.}
\label{fig:supp-davis-qualitative}
\end{figure}

\begin{figure}[!t]
    \centering
    \includegraphics[width=\textwidth]{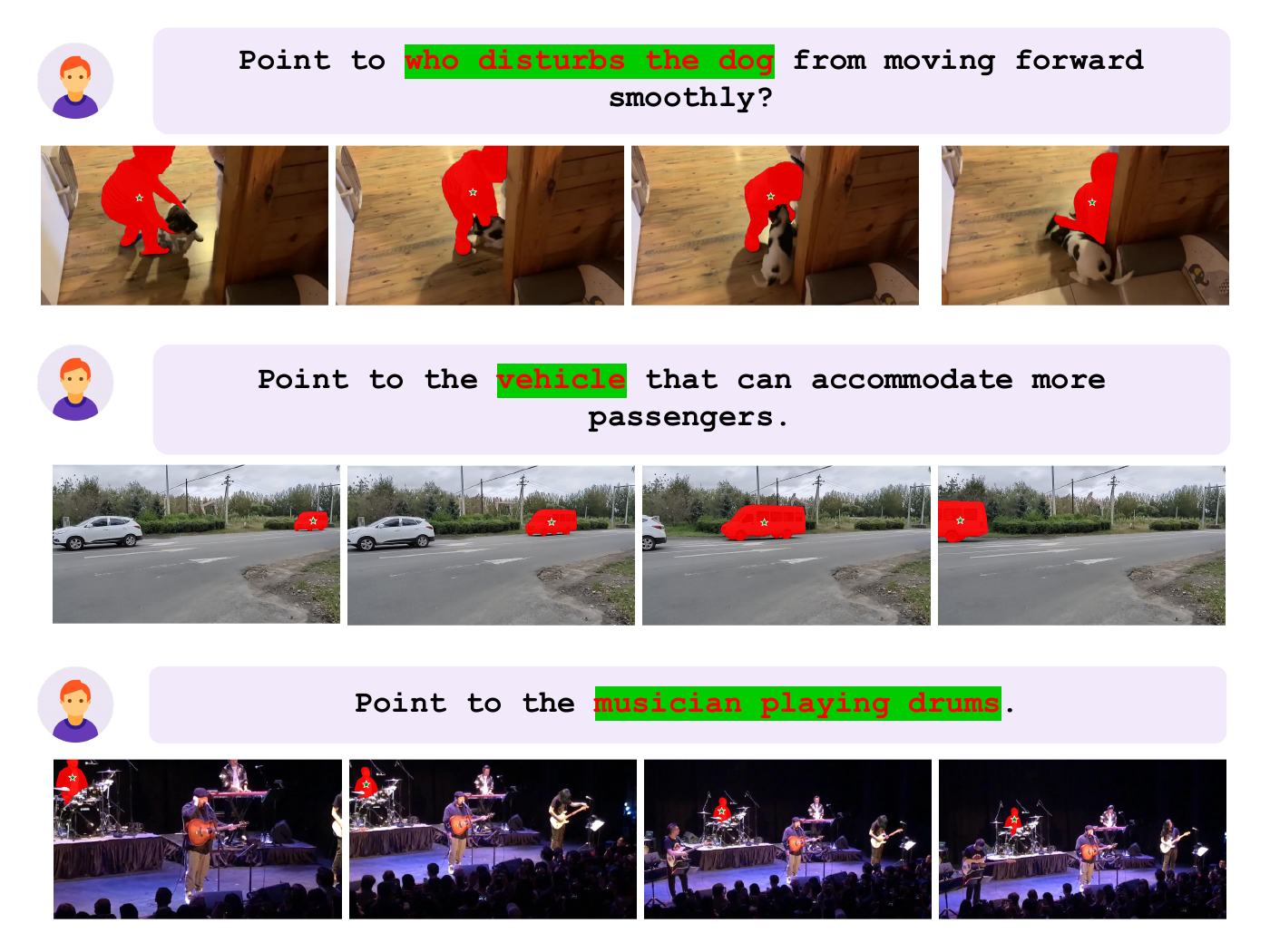}
    \caption{ReasonVOS qualitative examples.}
\label{fig:supp-reasonvos-qualitative}
\end{figure}

\textbf{General qualitative results. } 
We also present some qualitative results in Figures \ref{fig:supp-vpos-qualitative},\ref{fig:supp-mevis-qualitative},\ref{fig:supp-ytvos-qualitative}, \ref{fig:supp-davis-qualitative}, and \ref{fig:supp-reasonvos-qualitative} on our proposed VPoS-Bench, MeViS, YT-VOS, Ref-DAVIS, and ReasonVOS, respectively. We observe that in each case, VideoMolmo generates fine-grained points and corresponding masks pertaining to the query objects. In fact, VideoMolmo performs well even in the cases of multi-object queries (>2 objects) such as in VPoS-Bench counting task of Fig. \ref{fig:supp-vpos-qualitative} (1\textsuperscript{st} row) and Fig. \ref{fig:supp-mevis-qualitative} (3\textsuperscript{rd} row). Further, VideoMolmo also excels at grounding small and fine-grained objects. Fig. \ref{fig:supp-vpos-qualitative} (3\textsuperscript{rd} row) shows VideoMolmo accurately points and grounds the far-away car on the road, although the car is too small to point at in some frames. Similarly, VideoMolmo is able to ground the helmet in Fig. \ref{fig:supp-davis-qualitative} (2\textsuperscript{nd} row) while avoiding to ground the entire biker. 

\begin{figure}[!t]
    \centering
    \includegraphics[width=\textwidth]{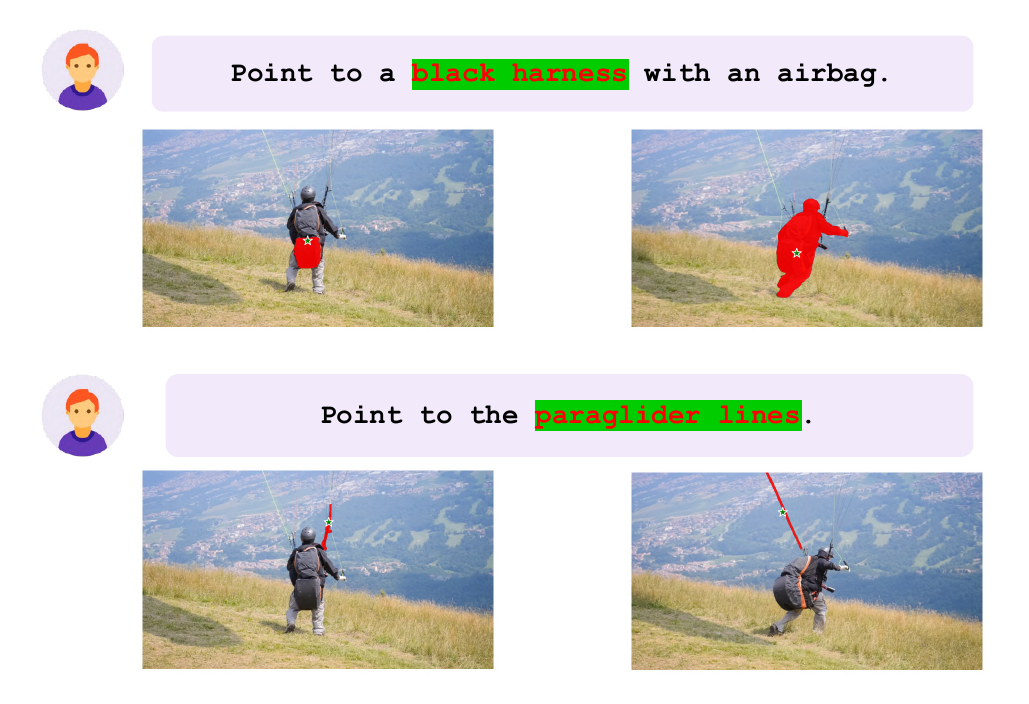}
    \caption{Qualitative failure cases of \method.}
\label{fig:supp-videomolmo_failure}
\end{figure}

\textbf{Failure cases. }
While VideoMolmo demonstrates strong fine-grained pointing capabilities, it is not without limitations. As illustrated in Fig.~\ref{fig:supp-videomolmo_failure}, certain failure cases highlight areas for improvement. In the first row, the model is expected to point to the black harness but instead grounds a part of the adjacent bag. This misalignment stems from the limitations of SAM2, which struggles to accurately convert the predicted point coordinate into a meaningful mask.
In the second row, the model points to only one of several visible paraglider lines, missing multiple lines. Such cases suggest a need for enhanced expressiveness, such as enabling the model to predict multiple points for a single query. Addressing these limitations opens new avenues for future work in improving the robustness and granularity of point grounding in complex scenes. 

%% file: tables/videomolmo_variants.tex
\begin{table}[h]
\centering
\caption{Performance comparison of different \method variants on Refer-DAVIS benchmark.}
\vspace{0.5em}
\begin{tabular}{l|c|c|c}
\hline
\textbf{Variant} & \textbf{$J$} & \textbf{$F$} & \textbf{$J\&F$} \\
\hline
VideoMolmo-O-7B & 66.25 & 72.93 & 69.59 \\
VideoMolmo (LoRA) & 67.82 & 74.72 & 71.27\\
VideoMolmo ($16\text{bit}$) & 67.74 & 74.72 & 71.25  \\
\hline
 \rowcolor{violet!10}
    VideoMolmo & \textbf{71.27} & \textbf{73.63} & \textbf{72.45} \\
\hline
\end{tabular}

\label{tab:videomolmo_variants}
\end{table}

%% file: tables/molmo_sampling_rate.tex
\begin{table}[t]
\centering
\setlength{\tabcolsep}{7pt}
\caption{Effect of sampling rate $k$ on Refer-DAVIS-17, Refer-Youtube-VOS, and MeViS benchmarks using Molmo + SAM2}
\resizebox{0.9\textwidth}{!}{
\begin{tabular}{l|ccc|ccc|ccc}
    \toprule
    \textbf{$k$} 
    & \multicolumn{3}{c|}{\textbf{Refer-DAVIS-17}} 
    & \multicolumn{3}{c|}{\textbf{Refer-Youtube-VOS}} 
    & \multicolumn{3}{c}{\textbf{MeViS}} \\
    \cmidrule(lr){2-4} \cmidrule(lr){5-7} \cmidrule(lr){8-10}
    & $\mathcal{J}$ & $\mathcal{F}$ & $\mathcal{J\&F}$ 
    & $\mathcal{J}$ & $\mathcal{F}$ & $\mathcal{J\&F}$ 
    & $\mathcal{J}$ & $\mathcal{F}$ & $\mathcal{J\&F}$ \\
    \midrule
    1        & 58.32 & 64.60 & 61.46 & 60.08 & 64.69 & 62.38 &  45.53  & 51.06   & 48.30   \\
    3        & 58.77 & 63.85 & 61.31 & 60.11 & 65.04 & 62.58 & --   & --   & --   \\
    10 & 59.61 & 64.77 & 62.19 & 60.24 & 64.88 & 62.56 & 45.63   & 50.93   & 48.28   \\
    30 & 63.31 & 69.31 & 66.31 & 59.48 & 64.02 & 61.75 & 45.51   & 50.65   & 48.08   \\
    $|\mathcal{T}|$ & 65.29 & 70.09 & 67.69 & 59.48 & 64.07 & 61.78 & 44.37 & 49.37 & 46.87   \\

    \bottomrule
\end{tabular}
}
\label{table:ablation-molmo-sam2-}
\vspace{-1em}
\end{table}

%% file: tables/threshold_ref_ytvos.tex
\begin{table}[!t]
\centering
\small
\setlength{\tabcolsep}{8pt}

\begin{minipage}{0.42\textwidth}
    \centering
    \caption{Effect of varying threshold $\tau$ on the performance of \method evaluated on the Refer-DAVIS benchmark.}
    \resizebox{\textwidth}{!}{
    \begin{tabular}{lccc}
    \toprule
    \textbf{$\tau$} & $\mathcal{J}$ & $\mathcal{F}$ & $\mathcal{J\&F}$ \\ 
    \midrule
    0   & 67.31 & 73.38 & 70.34 \\
    0.3   & 69.05 & 75.51 & 72.28 \\
    0.5   & 68.90 & 75.26 & 72.08 \\
    0.9   & 68.85 & 75.24 & 72.05 \\
    \hline
    \rowcolor{violet!10}
    VideoMolmo $(\tau = 0.7)$ & \textbf{71.27} & \textbf{73.63} & \textbf{72.45} \\
    \bottomrule
    \end{tabular}
    }
    \label{table:ablation-threshold-ref-davis}
\end{minipage}
\hfill
\begin{minipage}{0.48\textwidth}
    \centering \renewcommand{\arraystretch}{1.15}
    \caption{Effect of sampling rate
    $k$ on the performance of \method evaluated on the Refer-YouTube-VOS benchmark.}
    \resizebox{\textwidth}{!}{
    \begin{tabular}{lccc}
    \toprule
    \textbf{$k$} & $\mathcal{J}$ & $\mathcal{F}$ & $\mathcal{J\&F}$ \\ 
    \midrule
    3   & 64.39 & 67.68 & 66.03 \\
    10       & 65.69 & 69.39 & 67.54 \\
    15  & 66.26 & 69.95 & 68.11 \\
    20  & 66.34 & 69.93 & 68.14 \\
    30  & 65.80 & 69.32 & 67.56 \\
    \hline
    \rowcolor{violet!10}
    VideoMolmo ($k=5$) & \textbf{65.55} & \textbf{69.11} & \textbf{67.33} \\
    \bottomrule
    \end{tabular}
    }\renewcommand{\arraystretch}{1.0}

    \label{table:ablation-sampling_rate-ref-yt-vos}
\end{minipage}
\vspace{-1.6em}
\end{table}